# HEAD AUTOMATA FOR SPEECH TRANSLATION

*Hiyan Alshawi*

AT&T Research
600 Mountain Avenue
Murray Hill, NJ 07974, USA
hiyan@research.att.com

## ABSTRACT

This paper presents statistical language and translation models based on collections of small finite state machines we call "head automata". The models are intended to capture the lexical sensitivity of N-gram models and direct statistical translation models, while at the same time taking account of the hierarchical phrasal structure of language. Two types of head automata are defined: relational head automata suitable for translation by transfer of dependency trees, and head transducers suitable for direct recursive lexical translation.

## 1. INTRODUCTION

Speech translation demands robust and efficient models for translation. Our approach to satisfying these requirement is to provide language and translation models that are lexical and statistical, while still capable of capturing the hierarchical structuring of language into phrases.

Statistical approaches to translation are by their nature more robust than constraint-based approaches (for example [6], [7]) because the availability of a numerical function for comparing hypotheses allows the selection of the best translation from alternatives produced by models that overgenerate. In contrast, purely constraint-based translation systems are forced to limit the hypotheses produced at the expense of robustness in cases when none are produced. Statistical models can also lead to efficiency since the availability of a function for ranking partial hypotheses often makes efficient dynamic programming algorithms possible, as is the case for the statistical head automaton models described here.

Lexical models directly enhance efficiency in that only elements of the model relevant to lexical items in the input need to be considered. For example, in head automaton models, only the automata associated in the lexicon with words in the input come into play in analysis and translation. Unlike traditional stochastic context free grammars [2], lexical models facilitate training statistical models that are sensitive to lexical collocations and the idiosyncrasies of lexical items.

Of course, statistical-lexical language and translation models have existed for some time, notably N-gram models and the approach to statistical translation developed at IBM [3]. However, a characteristic of head automaton models is that they capture the phrasal structure of natural language because derivations in these models correspond to dependency trees. In dependency grammar (see, for example, [4]), a lexical item $w$ in a sentence is a *dependent* of some other lexical item, the *head* of $w$, resulting in a hierarchical phrase structure based entirely on lexical relations.

Head automata are small finite state machines associated with lexical items, or pairs of lexical items in the case of head transducers. In this paper, we describe two types of head automata. The first type, *relational head acceptors* (section 2) are the basis of monolingual language models suitable for analysis and generation in a transfer-based translator. The second type, head transducers (section 3) are the basis of bilingual models suitable for direct translation. We have implemented experimental speech translation systems using both model types, although we are only at the early stages of experimentation with head transducer models.

## 2. MONOLINGUAL HEAD AUTOMATA

### 2.1. Simple Head Acceptors

We first describe what is perhaps the simplest (monolingual) language model based on head automata. Relational head automaton models and bilingual head transducer models are elaborations of this model.

Each word $w$ in a vocabulary $V$ is associated with a *head acceptor* $M_w$. A head acceptor $M_w$ is a finite state machine with a set of states $Q$, one of which, $q_0$, is distinguished as the initial state. $M_w$ writes (or accepts) pairs $(L, R)$ of sequences of words from $V$, this being the characteristic difference between head acceptors and "standard" finite state acceptors. $M_w$ can undergo three types of action:

- *left transition*: if in state $q$, $M_w$ can write a word $w_l$ to the right end of $L$ with probability $P(w_l, \leftarrow, q'|M_w, q)$, and enter state $q'$.
- *right transition*: if in state $q$, $M_w$ can write a word $w_r$ to the left end of $R$ with probability $P(w_r, \rightarrow, q'|M_w, q)$, and enter state $q'$.
- *stop*: if in state $q$, $M_w$ can stop with probability $P(stop|M_w, q)$.

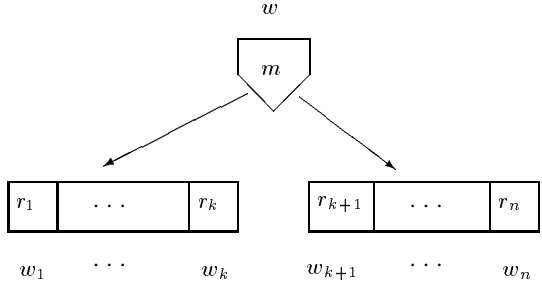

**Figure 1:** Head acceptor $m$ scans left and right sequences of relations $r_i$ for dependents $w_i$ of $w$.

For a consistent probabilistic model, the probabilities of all transitions and stop actions from a state $q$ must sum to unity. In a derivation of $L$ and $R$ by $M_w$, these sequences are initially empty and are considered complete when the machine undergoes a stop action.

When applied to language models, $L$ and $R$ are interpreted, respectively, as the sequences of dependent words of $w$ to its left and right in a string accepted by the model. The generative statistical model using simple head acceptors starts by choosing a head word $w_0$ for the entire string with probability $P(w_0|Top)$, followed by generating the sequences of dependents of $w_0$ according to $M_{w_0}$, generating the dependents of these dependents, and so on recursively. This yields an ordered tree structure, each local tree having been generated by the head acceptor for the root of the local tree. Leaves of the tree arise when a head acceptor undergoes a stop action in the initial state before making any transitions. A recursive left-parent-right traversal of the nodes of an ordered dependency tree for a derivation yields the word string for the derivation.

Head acceptors are formally more powerful than finite state automata that accept regular languages in the following sense. Each head acceptor defines a formal language whose strings are the concatenation of the left and right sequence pairs written by the automaton. The class of languages defined in this way clearly includes all regular languages, since strings of a regular language can be generated, for example, by a head acceptor that only writes a left sequence. Head acceptors can also recognize some non-regular languages requiring coordination of the left and right sequences, for example the language $a^n b^n$ (requiring two states), and the language of palindromes over a finite alphabet. Applying a cascade of head acceptors recursively as described above has the same generative power as context free grammars.

## 2.2. Relational Head Acceptors

A relational head acceptor is the same as a simple head acceptor described above except that the symbols written by transitions are not words but are taken from a set $K$ of (dependency) relation symbols. A relational head acceptor is thus a finite state machine that accepts (or writes) pairs of sequences of relation symbols. These correspond to the relations between a head word and the sequences of dependent phrases to its left and right (see Figure 1). Any state of a head acceptor can be an initial state, the probability of a particular initial state in a derivation being specified by a lexicon parameter

(explained below). The probability, given an initial state $q$, that automaton $m$ will a generate a pair of sequences, i.e.

$$P(\langle r_1 \cdots r_k \rangle, \langle r_{k+1} \cdots r_n \rangle | m, q)$$

is the product of the probabilities of the actions taken to generate the sequences.

## 2.3. Relational Head Acceptor Models

Each derivation in the generative statistical model based on relational head acceptors produces an *ordered dependency tree*, that is, a tree in which nodes dominate ordered sequences of left and right subtrees and in which the nodes have labels taken from $V$ and the arcs have labels taken from the set of dependency relations $K$. When a node with label $w$ immediately dominates a node with label $w'$ via an arc with label $r$, we say that $w'$ is an *r-dependent* of the *head* $w$. The model parameters for relational head acceptors (i.e. transition and stop action probabilities), together with the following dependency parameters and lexicon parameters, give a probability distribution for derivations.

A *dependency parameter* $P(\downarrow, w'|w, r')$ is the probability, given a head $w$ with a dependent arc with label $r'$, that $w'$ is the $r'$-dependent for this arc. A *lexicon parameter* $P(m, q|r, \downarrow, w)$ is the probability that a local tree immediately dominated by an $r$-dependent $w$ is derived by starting in state $q$ of automaton $m$. The model also includes lexicon parameters $P(w, m, q|Top)$ for the probability that $w$ is the head word for an entire derivation initiated from state $q$ of automaton $m$.

Let the probability of generating an ordered dependency subtree $D$ headed by an $r$-dependent word $w$ be $P(D|w, r)$. The recursive process of generating this subtree proceeds as follows:

1. Select an initial state $q$ of an automaton $m$ for $w$ with probability $P(m, q|r, \downarrow, w)$.

2. Run the automaton $m_0$ with initial state $q$ to generate a pair of relation sequences with probability $P(\langle r_1 \cdots r_k \rangle, \langle r_{k+1} \cdots r_n \rangle | m, q)$.

3. For each relation $r_i$ in these sequences, select a dependent word $w_i$ with dependency probability $P(\downarrow, w_i|w, r_i)$.

4. For each dependent $w_i$, recursively generate a subtree with probability $P(D_i|w_i, r_i)$.

We can now express the probability $P(D_0)$ for an entire ordered dependency tree derivation $D_0$ headed by a word $w_0$ as

$$\begin{aligned} P(D_0) = & \\ & P(w_0, m_0, q_0|Top) \\ & P(\langle r_1 \cdots r_k \rangle, \langle r_{k+1} \cdots r_n \rangle | m_0, q_0) \\ & \prod_{1 \leq i \leq n} P(\downarrow, w_i|w_0, r_i) P(D_i|w_i, r_i). \end{aligned}$$

In our speech translation application, we search for the highest probability derivation, rather than summing over all derivations yielding a string.

In practice, the number of parameters in a relational head acceptor language model is dominated by the dependency parameters, that is, $O(|V|^2|R|)$ parameters. This puts the size of the model somewhere in between 2-gram and 3-gram models. The similarly motivated link grammar model [5] has $O(|V|^3)$ parameters. Link grammar models also differ from the models presented here in that they can derive linkage graphs (parses) that are not trees.

## 3. BILINGUAL HEAD AUTOMATA

We now describe head transducers which form the basis of a translator that simultaneously applies collocations between words in each language, as well as word to word correspondences between the two languages. Models based on restricted head transducers are simple enough that they can be trained automatically from data consisting of example translations.

Collections of head transducers are used to convert (in either direction) between strings in a language with vocabulary $V_1$ and strings in a language with vocabulary $V_2$. We call the conversion process *recursive head transduction*. Unlike standard finite state transducers, translations of words in the input and output strings can be arbitrarily far apart without a corresponding increase in the number of model states. This is not possible for translation using standard transducers (for example, [8]).

A translator using head transducers consists of the following components:

- a collection of head transducers;
- a bilingual lexicon associating pairings of words (or phrases) from the two languages with particular head transducers;
- a parameter table specifying costs (typically negated log probabilities) for the actions taken by the transducers;
- a transduction search engine for finding the lowest cost translation of an input phrase or sentence.

The first three are described here. The transduction search engine uses a polynomial time dynamic programming algorithm similar in operation to the analysis algorithm for relational head acceptors [1].

### 3.1. Head Transducers

In a translator based on a collection of head transducers, each head transducer converts a pair of word sequences, the left and right dependents of a word $w \in V_1$ into another pair of word sequences, the left and right dependents of a word $v \in V_2$. A head transducer is thus a finite state machine for converting a pair of input sequences into a pair of output sequences (Figure 2). By applying a collection of head transducers recursively in a way analogous to the recursive application of simple head acceptors, we can convert the source language string into the target language string.

We augment the vocabulary $V$ of each language with the empty "word" $\epsilon$ to form $V' = V \cup \{\epsilon\}$. A head transducer $M$ consists of a set $Q$ of machine states, an initial state $q_0 \in Q$, and an action table. The machine reads a pair $(L_1, R_1)$ of $V_1'$ word sequences and writes a pair $(L_2, R_2)$ of $V_2'$ word sequences.

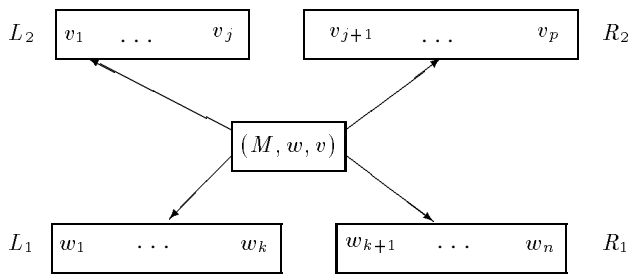

**Figure 2:** Head transducer $M$ converts the sequences of left and right dependents $\langle w_1 \ldots w_k \rangle$ and $\langle w_{k+1} \ldots w_n \rangle$ of $w$ into left and right dependents $\langle v_1 \ldots v_j \rangle$ and $\langle v_{j+1} \ldots v_p \rangle$ of $v$.

Each transition specifies a pair of *valencies* $(\alpha_1, \alpha_2)$, a valency being either $\leftarrow$ (left sequence) or $\rightarrow$ (right sequence), corresponding to the directions for transitions of simple head automata. $\alpha_1$ is applied to the source language and $\alpha_2$ to the target. For example, a transition with $(\leftarrow, \rightarrow)$ means read rightwards in $L_1$ and write to the left end of $R_2$. In terms of these valencies, the pair of valencies for transitions in a "standard" finite state transducer is always $(\leftarrow, \leftarrow)$.

To obtain a generative statistical model based on head transducers, we view a transducer as simultaneously deriving all four sequences of dependents. Specifically, the possible actions of a transducer $M$ in probabilistic derivations of the dependents of $w$ and $v$ are as follows:

- Stop: if in state $q$, transducer $M$ can stop with probability $P(stop|M, q)$, at which point the sequences are considered complete.
- Transition: if in state $q$, transducer $M$ can, with probability $P(w', v', \alpha_1, \alpha_2, q'|M, w, v, q)$, enter state $q'$ after writing a word $w' \in V_1'$ according to valency $\alpha_1$ and writing a word $v' \in V_2'$ according to valency $\alpha_2$.

The valencies $\{\leftarrow, \rightarrow\}$ are sufficient for most cases encountered in language translation and allow efficient polynomial transduction algorithms. A more general type of head transducer includes two additional valencies, $\Rightarrow$ and $\Leftarrow$, corresponding to writing or reading from the other ends of the sequences. For example, $\alpha_2 = \Rightarrow$ specifies writing to the *right* end of $R_2$.

### 3.2. Recursive Head Transduction

Recursive head transduction can be regarded as a probabilistic generative model for deriving pairs of strings, one in the source language the other in the target language. In addition to probability parameters associated with actions of the head transducers, the recursive head transduction generative model includes bilingual lexicon probability parameters: $P(M|w, v)$, the probability that $M$ is used to derive the dependents of $w$ and $v$; and $P(w_0, v_0|Top)$, the probability of $w_0$ and $v_0$ starting the derivation.

The derivation proceeds by the choice of starting words $w_0$ and $v_0$ with probability $P(w_0, v_0|Top)$, followed by selection of an initial head transducer $M_0$ to produce the dependent sequences of $w_0$ and $v_0$. For each transition with dependents $w'$ and $v'$, a transducer $M'$ is selected with probability $P(M'|w', v')$ to generate the dependent sequences of $w'$ and $v'$, and so on recursively.

A derivation in this model implicitly corresponds to building a pair of trees, $S$ and $T$, with ordered sequences of left and right children. $S$ has node labels taken from $V'_1$ and $T$ has node labels taken from $V'_2$. A recursive traversal of the nodes of $S$ and $T$ in left-parent-right order yields the source and target strings for the derivation.

A relational head transduction model can also be constructed in an analogous way to the relational head acceptor model.

### 3.3. A Trainable Transduction Model

We now consider a constrained case of recursive head transduction that can be used to train a translation model from a collection of translation examples. If a standard bilingual dictionary is available that simply specifies possible word translation pairs $(w, v)$ between the two languages, the parameter space for the constrained model is small enough for training large vocabulary translation models. Despite its simplicity, the parameters of the constrained model still simultaneously captures language to language lexical correspondences as well as monolingual word collocations in each of the two languages.

In the constrained model, all head transducers have a single state $q$ and the transition probability parameters are of the form

$$P(w', v', \alpha_1, \alpha_2 | w, v)$$

where $w \in V_1$, $v \in V_2$, $w' \in V'_1$ and $v' \in V'_2$; and $\alpha_1$ and $\alpha_2$ are valencies. Such a parameter corresponds to a $q$-to-$q$ transition of a head transducer $M$ for $(w, v)$ that reads $w'$ as specified by $\alpha_1$ and writes $v'$ as specified by $\alpha_2$.

A training procedure similar to the inside-outside algorithm can be used for estimating the model parameters by simultaneously analyzing both sentences of translation examples. If the model is constrained by a bilingual dictionary provided in advance, the number of parameters will be $O(|V_1|^2)$ under reasonable assumptions.

### 4. DISCUSSION

Head automata models are applicable to transfer-based translation and direct translation. In both cases, the models make use of collections of small finite state machines associated with entries in a monolingual or bilingual lexicon. The models are statistical and lexical, and well suited to bottom-up lattice processing, properties we believe to be important for efficient and robust speech translation.

We have implemented experimental speech translations systems using both types of head automata. These systems perform speaker independent English-Chinese translation in the ATIS domain whereby information requests made in English are translated to Chinese, and simple answers to these requests are translated from Chinese to English. So far, most of our effort has been concentrated on a transfer-based system in which the structure of the automata is specified manually followed by supervised and/or unsupervised parameter estimation. In this system, relational head acceptors are used for analysis of the source language and generation of the target language, while transfer is carried out according to a separate statistical tree-mapping model. The analysis, transfer, and generation algorithms, together with experimental evaluation of different parameter estimation methods are described in [1].

More recently, we have built a head-transducer version of the system in which approximate head transducer models were compiled from the models used in the transfer system. Head transducers simplify the overall design of the translation system but constrain the possible source to target mappings. It is not yet clear what practical effect this loss of generality has for translation applications. However, the additional constraints on mapping mean that fully automatic training of translation models is more feasible for head transducers. An additional benefit of head transducers is that they allow the target language to influence the analysis of the input since lexical associations in both languages are applied for each automaton transition, a property exploited by our current transduction search engine. This incremental application of the target language model may be important for future speech translation systems attempting simultaneous translation.